\documentclass{article}

\usepackage{arxiv}

\usepackage[utf8]{inputenc} 
\usepackage[T1]{fontenc}    
\usepackage{hyperref}       
\usepackage{url}            
\usepackage{booktabs}       
\usepackage{amsfonts}       
\usepackage{nicefrac}       
\usepackage{microtype}      
\usepackage{lipsum}		
\usepackage{multirow}
\usepackage{amsmath}
\usepackage{graphicx}
\usepackage{natbib}
\usepackage{doi}
\usepackage{xspace}
\usepackage{enumitem}
\usepackage{hyperref}
\usepackage{capt-of}
\usepackage{cleveref} 
\usepackage{booktabs}
\usepackage{pgfplots}
\pgfplotsset{compat=1.18}
\usepackage{tabularx}
\usepackage{xcolor}
\definecolor{softblue}{HTML}{1080E7}
\usepackage{algorithm}
\usepackage{algpseudocode}

\newcommand{\mypara}[1]{\noindent{\bf {#1}.}\xspace}
\newcommand{\systemname}{SAG-Mem\xspace}
\newcommand{\systemnamelite}{SAG-Mem-Lite\xspace}

\title{SAG-Mem-1.0 Technical Report}

\author{%
    \textbf{Yuchao Wu\textcolor{softblue}{\thanks{Correspondence to: \href{mailto:jomy@zleap.com}{\texttt{jomy@zleap.com}}.}}\kern0.35em, XingCheng Liang, Xuanze Wang, Yini Huang, Yue Wu}\\
    \textnormal{Zleap AI}\\
    \textnormal{\small\texttt{jomy@zleap.com, lensen@zleap.com, wangxuanze@zleap.com, huangyini@zleap.com, wuyue@zleap.com}}%
}

\date{}

\renewcommand{\headeright}{Zleap}
\renewcommand{\undertitle}{}

\hypersetup{
    colorlinks=true,
    citecolor=softblue,
    linkcolor=softblue,
    urlcolor=softblue
}

\begin{document}
\maketitle
\setcounter{footnote}{0}

\begin{abstract}
Long-term conversational memory requires retrieving complementary facts scattered across dialogue sessions, including intermediate evidence with limited similarity to the original question.
Existing approaches often rely on derived memory representations or persistent relational structures, introducing construction costs and potentially losing source details.
We present \systemname, a conversational memory framework that preserves complete dialogue messages in non-overlapping chunks and discovers complementary evidence through query-time lexical expansion.
Memory construction requires no model inference, using a lexical index without LLM-based extraction, summarization, or semantic relation construction.
At query time, BM25 retrieves initial candidates, an evidence evaluator identifies useful chunks, and lexical cues from selected evidence guide an additional retrieval round.
Retrieved dialogue is further annotated with calendar interpretations of supported relative-time expressions, preserving the original wording and temporal granularity.
We also introduce \systemnamelite, an alternative configuration that replaces evidence scoring with embedding-based candidate scoring, achieving lower operational cost than Full Context and all memory-based baselines across both benchmarks with low query latency.
On LoCoMo and LongMemEval, \systemname achieves overall scores of 83.08 and 80.20, respectively, the highest among the evaluated methods on both benchmarks.
On LongMemEval, it exceeds the strongest compared baseline by 7.20 points.
\systemnamelite achieves scores of 77.09 and 77.20, respectively.
These results support dialogue preservation and evidence-guided retrieval expansion as an effective design for long-term conversational memory.\footnote{Benchmark and evaluation code: \url{https://github.com/Zleap-AI/SAG-Mem-Benchmark}}
\end{abstract}


\enlargethispage{3\baselineskip}
\par\vfill
\noindent\begin{minipage}{\textwidth}
    \centering
    \includegraphics[width=0.95\linewidth]{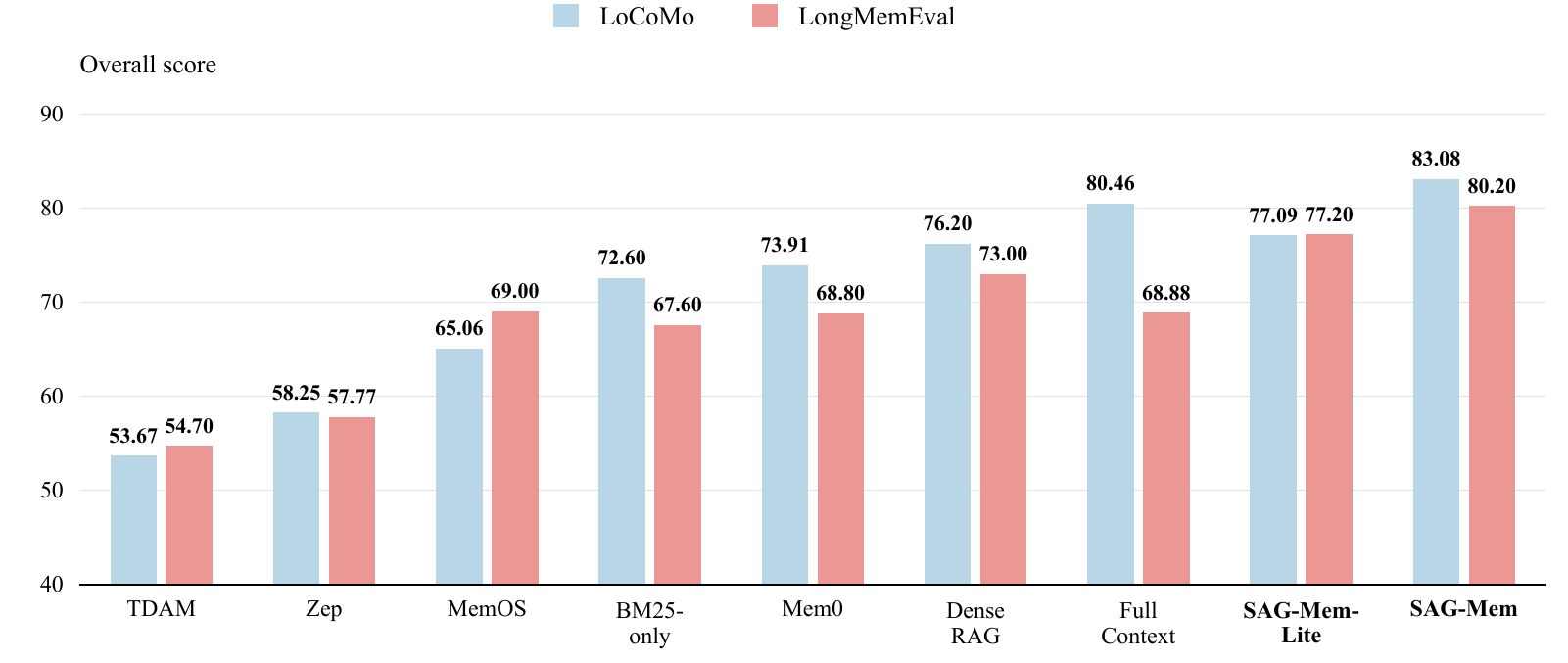}
    \captionof{figure}{
        Overall scores(\%) of different memory methods on LoCoMo and LongMemEval
        using Qwen3.6-35B as the backbone.
        SAG-Mem achieves the highest overall scores on both benchmarks, reaching 83.08\% on LoCoMo and 80.20\% on LongMemEval.
    }
    \label{fig:teaser}
\end{minipage}
\newpage

\noindent\begin{minipage}{\textwidth}
    \centering
    \includegraphics[width=\linewidth]{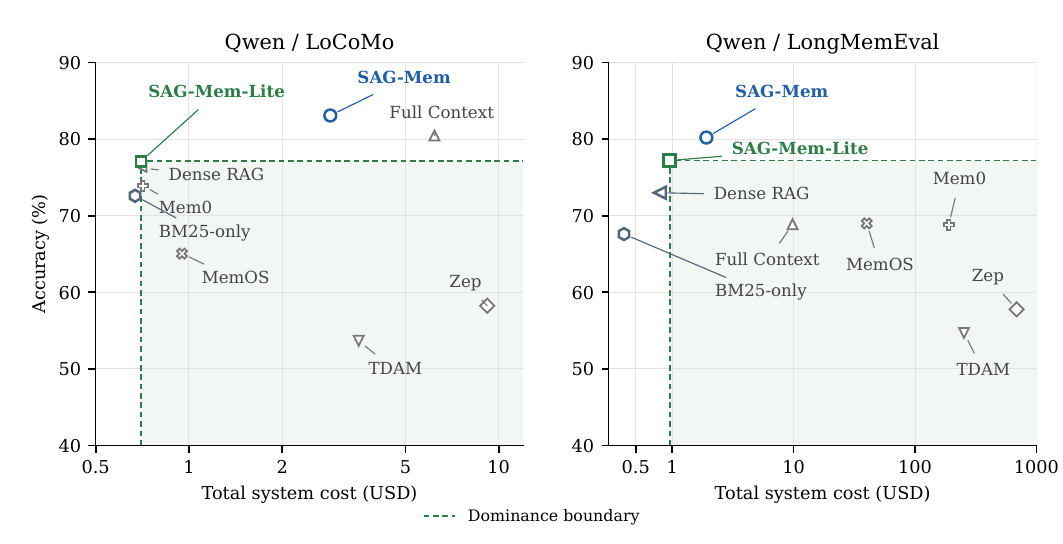}
    \captionof{figure}{Overall accuracy versus total operational cost under Qwen3.6-35B-A3B-FP8, using the values in Tables~\ref{tab:main_results} and~\ref{tab:cost}. The green dashed lines mark the cost and accuracy of \systemnamelite, and the shaded region highlights methods that cost more and achieve no higher accuracy. Both panels use logarithmic cost axes.}
    \label{fig:cost_accuracy}
\end{minipage}
\vspace{0.8em}

\section{Introduction}
\label{sec:introduction}

Long-term conversational memory enables language models to answer questions using information accumulated across past interactions~\citep{packer2023memgpt}.
As dialogue histories grow, retrieval-augmented generation (RAG) provides a way to access relevant information without placing the entire history in the model's context~\citep{lewis2020retrieval}.
However, answering a question may require complementary facts distributed across sessions, together with their conversational and temporal context~\citep{maharana2024evaluating,wu2025longmemeval}.
Retrieval based on similarity to the original question may miss intermediate evidence whose relevance emerges only after other facts are found~\citep{qi2019answering,trivedi2023interleaving}.
This motivates using retrieved evidence to guide further search beyond the initial candidate pool.

Memory systems address this problem through different choices of representation and organization.
MemoryBank~\citep{zhong2024memorybank} generates event summaries and user profiles, while RMM~\citep{tan2025prospect} summarizes interactions at multiple granularities.
Mem0~\citep{chhikara2025mem0} extracts and consolidates salient information into persistent memory entries.
For relational organization, A-MEM~\citep{xu2026mem} links memory notes and updates their contextual representations, while Zep~\citep{rasmussen2025zep} organizes extracted entities and facts in a temporal knowledge graph.
These representations can facilitate retrieval, but generating and updating them requires model inference and incurs recurring token costs during memory ingestion.
Moreover, details omitted from these representations may later be needed to answer a question or connect evidence across sessions.
Recent results with Raw-RAG~\citep{bao2026complex} show that dense retrieval and reranking over raw conversation history can outperform the compared agentic memory systems, highlighting the importance of preserving source information and the supporting relations needed to derive answers.
These observations motivate a design that stores source dialogue directly and uses retrieved evidence to guide further search for complementary information.

Temporal interpretation is another concern when using small Reader models~\citep{chu2024timebench,fatemi2024test}.
Expressions such as \textit{yesterday} and \textit{last month} depend on the date of the original message rather than the date of the current question~\citep{rasmussen2025zep}.
Even when the correct evidence is retrieved, the Reader must identify the appropriate reference date and translate the relative expression before answering.
Temporal normalization offers an established way to make such expressions explicit through standardized calendar representations~\citep{strotgen2010heideltime,chang2012sutime}.
This motivates providing explicit calendar interpretations of supported expressions to reduce the temporal reasoning required of the Reader.

These considerations lead to three research questions. \\
\textbf{RQ1: Memory construction.}
Can direct dialogue storage reduce costs while preserving the original details needed for accurate question answering across long conversations spanning multiple sessions? \\
\textbf{RQ2: Evidence discovery.}
Can evidence-guided lexical expansion recover complementary facts across sessions, including intermediate evidence that is poorly matched to the original query but needed to derive the answer? \\
\textbf{RQ3: Temporal interpretation.}
Can deterministic date annotations improve temporal question answering with small Reader models by resolving relative expressions against source message dates? 

Building on SAG2~\citep{wu2026sag2}, we present \textbf{\systemname}, which adapts evidence-guided lexical retrieval to conversational memory.
It indexes complete messages in non-overlapping chunks within each session, preserving speaker, date, and source identifiers without model inference.
At query time, BM25~\citep{robertson2009bm25} retrieves candidates, and JEV~\citep{almeida2026jev} scores their contribution to answering the question.
Terms selected deterministically from high-scoring chunks drive one additional BM25 search, followed by JEV rescoring of retained and new candidates in a shared evidence context.

\textbf{\systemnamelite} shares the same chunks and lexical expansion, replacing JEV with query--chunk cosine similarity computed using bge-large-en-v1.5~\citep{xiao2024c}.
It precomputes one embedding per chunk and supports up to two expansion rounds.
Both configurations return source dialogue with date-based annotations of relative-time expressions.
The Reader is instructed to answer from this evidence and abstain when it is insufficient.

The key contributions of \systemname and \systemnamelite can be summarized as follows:
\begin{itemize}[leftmargin=12pt, itemsep=2pt, topsep=0pt, parsep=0pt, partopsep=0pt]
    \item We present a dialogue-preserving memory design that avoids LLM-based extraction and persistent semantic relation construction, together with date-based annotations that make supported relative-time expressions explicit while preserving the original text and temporal granularity.
    \item We develop seed-based lexical expansion to discover complementary evidence beyond the initial retrieval pool, using JEV-based evidence evaluation in \systemname and embedding-based candidate scoring in \systemnamelite.
    \item We evaluate both configurations on LoCoMo and LongMemEval. As shown in \Cref{fig:teaser}, \systemname achieves overall scores of 83.08 and 80.20, respectively, the highest among the evaluated methods on both benchmarks, while \systemnamelite achieves 77.09 and 77.20.
\end{itemize}

\section{Problem Statement}
\label{sec:problem-statement}

We study long-term conversational question answering over a history of past interactions.
Let $\mathcal{H}=(S_1,\ldots,S_N)$ denote a history comprising $N$ chronologically ordered sessions, where each session $S_i=(m_{i,1},\ldots,m_{i,n_i})$ contains a sequence of messages with associated date information.
Each message has its original text, speaker, and source identifiers.
Given a question $q$ about $\mathcal{H}$, the task is to produce an answer supported by the recorded interactions.
The required evidence may span multiple sessions, involve intermediate facts, and depend on the temporal context of the source messages.

In our setting, each memory record is a dialogue chunk containing consecutive complete messages from a single session.
These chunks are non-overlapping and retain the original message text and metadata.
Let $\mathcal{C}$ denote the resulting collection of source chunks.
A memory system constructs a memory store $\mathcal{M}=\operatorname{Build}(\mathcal{H})$ from these chunks.
Given an evidence budget $K$, the system retrieves an ordered list of chunks from $\mathcal{C}$:
\begin{equation}
    \mathcal{E}_q=\operatorname{Retrieve}(q,\mathcal{M}),
    \qquad |\mathcal{E}_q|\leq K,
    \label{eq:memory-retrieval-task}
\end{equation}
where $K$ limits the number of chunks passed to the Reader.

Let $g(\mathcal{E}_q)$ render the selected source dialogue together with its speaker, date, session, and message metadata.
This rendering also appends calendar interpretations of supported relative-time expressions, using the associated source-message dates as reference points while preserving the original wording and temporal granularity.
The Reader generates an answer as
\begin{equation}
    \hat{a}=\operatorname{Reader}\!\left(q,g(\mathcal{E}_q)\right).
    \label{eq:memory-answer-task}
\end{equation}

The retrieval objective is to recover complementary evidence that jointly supports answering $q$ within the budget $K$.
This includes intermediate facts that connect information across sessions or provide clues for locating additional evidence, even when they have limited direct similarity to the original question.
The answer should remain consistent with the source dialogue, relevant temporal conditions, and the precision supported by the evidence.
When the retrieved evidence is insufficient, including when the requested information is absent from $\mathcal{H}$, the expected response is to acknowledge that the available information is insufficient.


\section{\systemname Framework}
\label{sec:method}

Section~\ref{sec:introduction} motivated three design decisions: storing complete dialogue instead of extracted memories, discovering cross-session links at query time, and clarifying relative-time expressions before reading.
All three follow one principle: deferring each act of interpretation until the information it needs becomes available.
At write time the question is unknown, so the dialogue is stored verbatim.
At query time the evidence retrieved so far indicates what to look for next, so links between facts are discovered then.
At read time it is known which messages the Reader will see, so relative times are resolved then, for those messages only.

\begin{figure*}[!t]
\centering
\includegraphics[width=\textwidth]{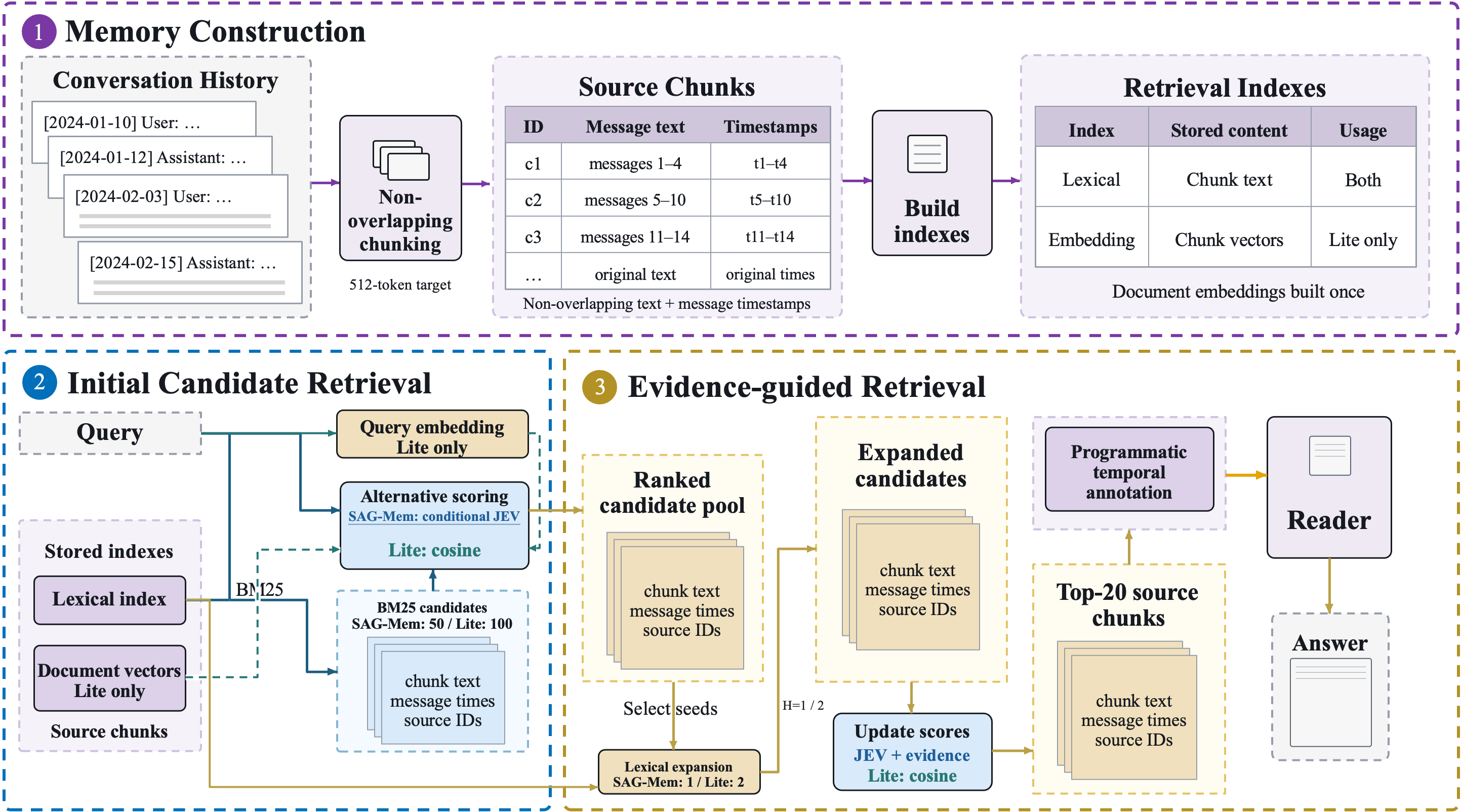}
\caption{Overview of the pipeline. Complete messages are grouped into source-linked chunks, retrieved evidence guides a further lexical search, and selected messages receive deterministic time clarifications before answer generation. Elements marked ``Lite'' belong to the variant described in Section~\ref{sec:lite}. The number of messages per chunk is illustrative; chunk length follows the 512-token budget of Section~\ref{sec:construction}.}
\label{fig:framework}
\end{figure*}

This section instantiates the three components of \Cref{sec:problem-statement} in pipeline order (\Cref{fig:framework}).
Section~\ref{sec:construction} gives $\operatorname{Build}$, Sections~\ref{sec:initial} and~\ref{sec:expansion} give $\operatorname{Retrieve}$ of Eq.~\eqref{eq:memory-retrieval-task}, and Section~\ref{sec:answer} gives the rendering $g$ and the Reader of Eq.~\eqref{eq:memory-answer-task}.
Table~\ref{tab:challenges} summarizes which concern each stage addresses.
Section~\ref{sec:lite} then presents \systemnamelite, which keeps the store and the retrieval procedure but replaces the scoring model and changes the search budgets.

\begin{table}[t]
\centering
\caption{How each stage addresses a concern raised in Section~\ref{sec:introduction}.}
\label{tab:challenges}
\begin{tabularx}{\linewidth}{
    @{}l
    >{\raggedright\arraybackslash}p{0.25\linewidth}
    >{\raggedright\arraybackslash}X@{}
}
\toprule
Concern & Stage & Design choice \\
\midrule
RQ1 & Memory construction & Complete messages stored verbatim with append-only writes \\
RQ2 & Initial and evidence-guided retrieval & A second lexical search driven by terms from selected chunks, plus rescoring in the context of gathered evidence \\
RQ3 & Answer generation & Deterministic date annotations after retrieval while keeping source wording \\
\bottomrule
\end{tabularx}
\end{table}

\subsection{Memory Construction}
\label{sec:construction}

Memory construction implements $\operatorname{Build}$.
Extraction-based memories decide at write time which details to retain~\citep{zhong2024memorybank,chhikara2025mem0}.
Because the system cannot know then what a later question will need, \systemname makes no such judgment and stores the dialogue as it was said.

\mypara{Complete-Message Chunks}
We process $\mathcal{H}$ in source order and group consecutive messages into chunks, each lying within a single session and disjoint from the others.
Packing follows a soft budget of 512 tokens: if appending the next complete message would exceed the budget, the current chunk is closed first.
Token counts use the bge-large-en-v1.5 tokenizer~\citep{xiao2024c} and include rendered headers and special tokens.
The tokenizer serves only as a length measure for packing and takes no part in indexing or scoring.
Messages always stay whole, and any single message that exceeds 512 tokens forms its own chunk.
The retrieval unit is therefore always a span of original dialogue, a granularity choice that affects both retrieval and reading~\citep{pan2025secom}.

\mypara{Rendered Text and Provenance}
Each message is rendered with its speaker and source date, so both become part of the chunk text.
\systemname stores no derived statements: a memory record in the sense of Section~\ref{sec:problem-statement} is simply such a chunk together with its provenance,
\begin{equation}
c_i=(x_i,\mathcal{R}_i),\qquad c_i\in\mathcal{C},
\label{eq:record}
\end{equation}
where $x_i$ is the rendered source text and $\mathcal{R}_i$ holds the session identifier, the message identifiers, and the full message span that the chunk covers.

\mypara{Source Dates}
A source date records when a message was said, whereas the event it describes may have happened at another time.
A message keeps its own date when one is present, and inherits the session date otherwise.
Relative-time expressions stay verbatim in storage.
The date they depend on is stored beside them, so resolving them can wait until after retrieval (Section~\ref{sec:answer}).

\mypara{Append-Only Maintenance}
Chunks from a new session are appended to $\mathcal{M}$ and added to a lexical index over normalized text.
A correction and the statement it revises coexist, each with its own source and date.
The store neither merges them nor assigns validity intervals, and whichever of them are retrieved reach the Reader as dated source text.
For \systemname, construction as a whole consists of tokenization and indexing and involves no model inference; the Lite variant adds one embedding per chunk on top of it (Section~\ref{sec:lite}).


\subsection{Initial Candidate Retrieval}
\label{sec:initial}

Construction produces only verbatim chunks and a lexical index over them, so connecting facts is the job of $\operatorname{Retrieve}$.
Retrieval follows SAG2~\citep{wu2026sag2}, an embedding-free multi-hop retriever over a lexical index.
We adopt its BM25 candidate pool, its evidence scoring with JEV, its selection of the highest-scoring chunks as seeds, its weighting of expansion terms, its per-term retrieval with interleaving, and its protection of the first five BM25 matches.
We change the retrieval unit to complete-message chunks and the budgets to $B{=}50$ and $H{=}1$, and we define the pool of every round, including the last, as the set $U_t$ of Section~\ref{sec:expansion}, without SAG2's refilling of the final pool with earlier candidates.
Retrieval proceeds in two steps.
This section finds chunks that overlap the question lexically and selects seeds among them, and Section~\ref{sec:expansion} then starts from the seeds to reach bridging chunks that the question alone cannot reach~\citep{xiong2019simple}.

\mypara{Lexical Candidate Pool}
We retrieve an initial pool $C_0$ of at most $B{=}50$ chunks from the lexical index with BM25~\citep{robertson2009bm25}.
$B$ bounds the number of chunks admitted by each lexical search, both this initial one and every expansion round of Section~\ref{sec:expansion}.
The query is the disjunction of the question terms after stop-word removal and normalization.
The searchable content is the chunk text, and source identifiers serve as metadata.

\mypara{Evidence Scoring}
BM25 rank reflects term overlap, whereas seed selection requires knowing whether a chunk helps answer the question.
We therefore rerank $C_0$ with the evidence evaluator JEV~\citep{almeida2026jev}.
JEV judges whether a candidate's own text supplies a fact required to answer $q$, including an intermediate fact that connects other evidence.
The initial score is
\begin{equation}
d_i^{(0)}=J_\theta\bigl(q,x_i;E_0,V_0(i)\bigr),\qquad E_0=\emptyset,
\label{eq:jev0}
\end{equation}
where $E_0$ is the explicit evidence list and $V_0(i)$ is the set of candidate passages visible in the same request as $c_i$.
No evidence has been selected yet, so $E_0$ is empty, although the other candidates in the request remain visible.
Appendix~\ref{app:packing} describes how a pool is packed into requests.

\mypara{Seeds}
The $s{=}10$ highest-scoring chunks form the set of expansion seeds $Z_0$.
Seeds serve two roles: they remain candidates for the final list, and they supply terms for the next search.

\subsection{Evidence-Guided Retrieval}
\label{sec:expansion}

The initial stage yields the seeds, which are chunks that overlap the question and received the highest evidence scores.
What is still missing are facts that share terms with a seed but few with the question itself.

\mypara{From the Question to the Evidence}
Consider the question ``Where does Maya's mentor live?''.
Session $S_1$ contains Maya's line ``My mentor is Leo.'', which overlaps the question and can become a seed.
Session $S_7$ contains Leo's line ``I live in Kyoto.'', which shares only the verb ``live'' with the question and mentions neither Maya nor a mentor, so a search issued with $q$ has little to match it on.
The seed, however, contains the term ``Leo'', and so does the rendered text of $S_7$ because the speaker name is part of the chunk.
Searching again with terms from the seed reaches the bridging fact.
A graph-based system would store this edge at write time~\citep{rasmussen2025zep,xu2026mem}, and iterative retrievers reach it by generating intermediate reasoning or reformulated queries~\citep{trivedi2023interleaving}.
Here it is recovered directly from the text when needed, which is what allows construction to stay lightweight.

\mypara{Expansion Terms}
Expansion terms are taken directly from the text of the seed chunks, so the step is deterministic and purely lexical.
Unlike generative query expansion~\citep{wang2023query2doc}, it needs no generative model.
It resembles pseudo-relevance feedback~\citep{rocchio1971relevance}, except that terms come only from the seeds, which are selected by the scoring function rather than by BM25 rank.
The weight of a term combines four signals, as defined in SAG2: its term frequency, its document frequency in the corpus, the overlap between its sentence and the question, and the score of its seed.
At most 20 terms are kept, and any term composed entirely of question words is dropped because the first search already used it.

\mypara{Collecting New Candidates}
Each expansion term retrieves matching chunks that no earlier search has returned, ranked by a query that combines the question terms with the expansion terms.
In the first round these are chunks outside $C_0$; in round $t$ of a multi-round run they are chunks outside $C_0\cup\cdots\cup C_{t-1}$, so no chunk is retrieved twice.
The per-term lists are interleaved by rank and deduplicated into at most $B$ new candidates $C_t$, which form $C_1$ in the first round.
Interleaving lets every term contribute, so that one frequent term cannot fill the budget.
If no usable term or no new candidate is found, the current ranking is kept and retrieval stops here.

\mypara{Rescoring with Evidence}
A new candidate enters the pool only because it shares a term with a seed, which does not make it useful.
It still has to be judged, and this judgment can now draw on the evidence found so far.
The second-round pool is $U_1=P\cup Z_0\cup C_1$, where $P$ contains the first five chunks of $C_0$ in BM25 order, and the members of $C_0$ outside $P\cup Z_0$ leave the pool.
The shared evidence context is $E_1=Z_0\cup\mathrm{First}_s(C_1)$, with new candidates in retrieval order, and JEV rescores every member of $U_1$:
\begin{equation}
d_i^{(1)}=J_\theta\bigl(q,x_i;E_1,V_1(i)\bigr),
\label{eq:jev1}
\end{equation}
where $V_1(i)$ is defined as $V_0(i)$ for the requests of this round.
$E_1$ is shared across all decisions of the round and may include the candidate being scored.
In the running example, ``I live in Kyoto.'' is judged alongside ``My mentor is Leo.'', making its role as an intermediate fact visible.

\mypara{Final Order}
The pool is ordered by the latest JEV score, with ties broken by retrieval rank and then by chunk identifier.
The first $K{=}20$ chunks are returned as $\mathcal{E}_q$ of Eq.~\eqref{eq:memory-retrieval-task}.
Algorithm~\ref{alg:retrieve} states the procedure for a general number of rounds $H$.
\systemname runs a single round, in which the seeds, new candidates, pool, and evidence of the algorithm are $Z_0$, $C_1$, $U_1$, and $E_1$ above.

\begin{algorithm}[t]
\caption{Evidence-guided retrieval, adapted from SAG2~\citep{wu2026sag2}. $\mathrm{Score}(c;E)$ is the scoring function evaluated with evidence list $E$. The pool of round $t$ is $U_t=P\cup Z_0\cup\cdots\cup Z_{t-1}\cup C_t$. $E_t$ is specified for $t{=}1$, the default setting of \systemname; \systemnamelite ignores the evidence list. \systemname uses JEV scores and $H{=}1$; Section~\ref{sec:lite} gives the settings of \systemnamelite.}
\label{alg:retrieve}
\begin{algorithmic}[1]
\Require question $q$, store $\mathcal{M}$, scoring function $\mathrm{Score}$, budgets $B$, $H$, $s$, $K$
\Ensure $\mathcal{E}_q$ with $|\mathcal{E}_q|\le K$
\State $C_0 \gets \mathrm{BM25}(q,\mathcal{M},B)$ \Comment{initial lexical pool}
\State score every $c\in C_0$ with $\mathrm{Score}(c;\emptyset)$
\State $P \gets$ first five chunks of $C_0$ in BM25 order; \ $U \gets C_0$
\State $Z_0 \gets$ top-$s$ chunks of $C_0$ by score \Comment{initial seeds}
\For{$t=1,\ldots,H$}
    \State $T \gets \mathrm{ExpandTerms}(Z_{t-1},q)$ \Comment{at most 20 terms}
    \State $C_t \gets \mathrm{LexicalExpand}\bigl(T,q,\mathcal{M}\setminus(C_0\cup\cdots\cup C_{t-1}),B\bigr)$ \Comment{unseen chunks only}
    \If{$T=\emptyset$ \textbf{or} $C_t=\emptyset$}
        \State \textbf{break} \Comment{keep the current ranking}
    \EndIf
    \State $U \gets P\cup Z_0\cup\cdots\cup Z_{t-1}\cup C_t$ \Comment{pool $U_t$}
    \State $E_t \gets$ evidence list of round $t$, with $E_1=Z_0\cup\mathrm{First}_s(C_1)$ \Comment{read only by JEV}
    \State score every $c\in U$ with $\mathrm{Score}(c;E_t)$
    \If{$t<H$}
        \State $Z_t \gets$ top-$s$ chunks of $U\setminus(Z_0\cup\cdots\cup Z_{t-1})$ by score \Comment{new seeds}
    \EndIf
\EndFor
\State order $U$ by score, then retrieval rank, then chunk identifier
\State \Return the first $K$ chunks of $U$ as $\mathcal{E}_q$
\end{algorithmic}
\end{algorithm}

\subsection{Answer Generation}
\label{sec:answer}

Answer generation implements the rendering $g$ and the Reader of Eq.~\eqref{eq:memory-answer-task}.
The retrieved chunks arrive in their original wording, and one act of interpretation remains before the Reader sees them: turning relative-time expressions into calendar dates.

\mypara{Time Clarification after Retrieval}
The selected chunks $\mathcal{E}_q$ are rendered in retrieval order with their complete messages, speakers, source dates, and session and message identifiers.
A deterministic renderer inserts a parenthetical date after each supported relative-time expression and keeps the original wording.
For a message dated 2025/05/12, ``yesterday'' becomes ``yesterday (2025/05/11)''.
The renderer consists of pattern matching and calendar arithmetic, following rule-based temporal taggers such as SUTime~\citep{chang2012sutime} and HeidelTime~\citep{strotgen2010heideltime}.
Month-level and year-level expressions keep their granularity, and unsupported expressions are left unchanged.

We apply this step after retrieval for two reasons.
First, the annotation is needed only for the messages the Reader will see, and each of them carries the date on which it was said, so a rule can take over a computation on which language models remain error-prone~\citep{fatemi2024test}.
Second, the annotation exists only in the Reader's context: the index, chunk identifiers, and retrieval scores remain unchanged, and the original wording stays available.

\mypara{Calendar Rules}
The renderer applies the following conventions.
Week-level expressions such as ``last week'' denote a calendar week that runs from Monday to Sunday.
Counted expressions such as ``two weeks ago'' are instead offsets in days, here 14 days, and do not denote a calendar-week interval.
Weekday expressions follow a separate rule: an English ``last'' or ``next'' weekday refers to the nearest matching day strictly before or strictly after the source date.
Approximate expressions stay approximate, and text inside quotations and code is left unchanged.
Dates are written in the form \texttt{2025/05/12}, and an inline annotation takes the form \texttt{yesterday (2025/05/11)}.
The rules cover English and Chinese expressions, although the dialogue in the experiments of this report is English.

\mypara{Reader}
The Reader receives $q$ and the rendered context $g(\mathcal{E}_q)$ and generates $\hat{a}$.
In line with Section~\ref{sec:problem-statement}, it is instructed to answer concisely and to abstain when the evidence is insufficient.
Its task is to interpret the source messages together with their dates, including references and updates across sessions.

\subsection{\systemnamelite}
\label{sec:lite}

In the pipeline above, the only retrieval step that spends model inference at query time is JEV scoring.
\systemnamelite keeps the store, lexical search, expansion procedure, and time rendering unchanged, replaces JEV with embedding similarity, and uses its own search budgets.
Its only model call during retrieval is a single embedding of the question.

\mypara{Cosine Scoring}
\systemnamelite encodes the question and the chunks with bge-large-en-v1.5~\citep{xiao2024c} into L2-normalized vectors.
Chunk vectors are computed once at build time, which is the only model cost that \systemnamelite adds to construction.
A vector is used only for scoring, and $x_i$ remains the text that is searched and read.
The question is encoded once at query time.
The score is
\begin{equation}
d_{\mathrm{Lite}}(c_i)=\bar{e}_q^{\top}\bar{e}_i,
\label{eq:lite}
\end{equation}
which equals cosine similarity because both vectors are normalized.

\mypara{Scoring without Evidence}
In Algorithm~\ref{alg:retrieve}, \systemnamelite sets $\mathrm{Score}$ to $d_{\mathrm{Lite}}$.
This score depends only on the fixed question vector and ignores the evidence list, so each chunk is scored once and rescoring reduces to scoring the new candidates.

\mypara{A Wider Search}
Candidates come from the same two sources as in \systemname, the initial BM25 search and lexical expansion.
Unlike in dense retrieval~\citep{karpukhin2020dense}, the embeddings here only rank these candidates, and no nearest-neighbor search runs over the full store.
Because cosine scoring is inexpensive, \systemnamelite spends the saved budget on a wider search: up to $B{=}100$ chunks per lexical search and up to $H{=}2$ expansion rounds, the default budgets of SAG2.
The weighting of expansion terms uses the score of the seed; since a cosine value can be negative, the score is mapped to $(d_{\mathrm{Lite}}+1)/2$ to keep it non-negative.
The mapping is monotone, so seeds keep their cosine order, and the mapped score enters the term weights in the same way as a JEV score.

\mypara{Pool across Rounds}
After the first round, up to $s$ further seeds are selected by cosine score among the chunks of the current pool that have not served as seeds before.
The pool of round $t$ is $U_t=P\cup Z_0\cup\cdots\cup Z_{t-1}\cup C_t$: it retains the first five initial BM25 matches and all earlier seeds alongside the new candidates of the round.
The pool is thus rebuilt in every round, and a candidate of an earlier round that is neither in $P$ nor chosen as a seed leaves it, as the members of $C_0$ outside $P\cup Z_0$ leave after the first round.
Its size in round $t$ is therefore at most $5+ts+B$, so the pool grows only by the $s$ seeds of each round rather than by all earlier candidates.
When the second expansion finds new candidates, a first-round candidate reaches the final list only by being selected as a seed, whereas every candidate of the second round competes for it directly; when it finds none, retrieval stops and $U_1$ is kept.
The final list is ordered by cosine similarity, and at most $K{=}20$ chunks are returned.

\mypara{Cost of the Substitution}
The cost of this substitution is that the scores of \systemnamelite no longer depend on the evidence gathered: a bridging chunk that looks useful only next to a seed is ranked by its similarity to the question alone.
The two variants can therefore rank the same candidates differently and, as a result, select different seeds.
With seeds chosen by question similarity alone, the expansion of \systemnamelite is closer to pseudo-relevance feedback over a reranked list.

In summary, the two variants share the store, the lexical search, the expansion procedure, and the time rendering, and differ in how candidates are scored and in the budgets $B$ and $H$.
A comparison between them is therefore a comparison of two configurations rather than of the scoring functions alone.
Section~\ref{sec:experiments} gives their configurations and evaluates both in terms of accuracy and cost.

\section{Experiment}
\label{sec:experiments}
\subsection{Experimental Setup}
\label{sec:setup}

\mypara{Dataset and Models}
We evaluate on the text-only setting of LoCoMo~\citep{maharana2024evaluating} (excluding image captions) and the LongMemEval-S benchmark~\citep{wu2025longmemeval}, which contain 1,986 and 500 questions, respectively. For both answer generation (Reader) and LLM-as-a-Judge (Judge)~\citep{zheng2023judging}, we use two configurations: \texttt{Qwen3.6-35B-A3B-FP8}~\citep{qwen36_35b_a3b} and \texttt{DeepSeek-V4.1-Flash-1017}~\citep{xu2026deepseek}. 
Both configurations disable thinking mode and provider fallback, using temperature~0.

\mypara{Baselines}
We compare \systemname and \systemnamelite against three groups of baselines: (1)~\textbf{Full Context}, which passes the complete conversation history directly to the Reader; (2)~\textbf{Memory-Based Systems}, including Zep~\citep{rasmussen2025zep}, Mem0~\citep{chhikara2025mem0}, TencentDB-Agent-Memory (TDAM)~\citep{tencentdb_agent_memory}, and MemOS~\citep{li2025memos}; and (3)~\textbf{Standard RAG Baselines} operating over raw conversation history, specifically \textbf{BM25-only} and \textbf{Dense RAG}. All baselines are reproduced under the same Reader and Judge configuration. \Cref{tab:versions} summarizes all software versions and model identifiers.

\mypara{Evaluation Metrics}
We evaluate performance along three dimensions:
\begin{itemize}[leftmargin=12pt, itemsep=2pt, topsep=0pt, parsep=0pt, partopsep=0pt]
    \item \textbf{Accuracy:} Binary answer correctness judged by LLM-as-a-Judge using dataset-specific semantic matching and abstention prompts, with rare API failures excluded from the denominator.
    \item \textbf{Retrieval Recall:} Session-level and turn-level Recall@$K$ for $K \in \{5, 10, 20\}$, evaluated against ground-truth evidence annotations over 1,973 annotated LoCoMo questions and all 500 LongMemEval questions.
    \item \textbf{Efficiency:} Monetary cost and end-to-end query latency, measuring query-time retrieval and Reader generation over prebuilt indexes while excluding offline indexing and judging.
\end{itemize}

\mypara{Default Configuration}
Memory construction and chunking follow the specification in \Cref{sec:method}. Across both benchmarks, \systemname retrieves up to $B=50$ candidates per round with at most $H=1$ lexical expansion round, whereas \systemnamelite retains the default SAG2 budgets of $B=100$ and up to $H=2$ rounds. Both variants select $s=10$ expansion seeds, retain up to 20 expansion terms, and pass the top $K=20$ chunks to the Reader. These default settings are held fixed across all main evaluations without per-dataset tuning; we evaluate retrieval parameter sensitivity around this default configuration in \Cref{sec:sensitivity}.

\mypara{Unified Cost Accounting}
Operational cost covers memory construction, query-time retrieval, and Reader answer generation, excluding offline LLM judging; construction is counted once per distinct conversation history. All systems are priced from logged token counts using official API pricing, accounting for prompt-cache hit discounts across extraction LLMs, search/Reader LLMs, evidence evaluators, and both construction-time and query-time embeddings. Reused retrieval logs contribute their full logical cost to each independent evaluation configuration.

\begin{table}[!htbp]
\centering
\small
\caption{Software versions and model identifiers used across all evaluated systems.}
\label{tab:versions}
\begin{tabular}{lll}
\toprule
\textbf{System / Role} & \textbf{Component} & \textbf{Version or Model Identifier} \\
\midrule
Zep & Memory backend & Graphiti \texttt{0.29.3} \\
Mem0 & Memory backend & Mem0 \texttt{2.2.1} \\
MemOS & Memory backend & MemOS \texttt{2.0.32} \\
TDAM & Memory backend & TencentDB-Agent-Memory (GitHub, Nov 2026) \\
BM25-only & Lexical retrieval & BM25~\citep{robertson2009bm25} \\
Dense RAG & Embedding model & \texttt{text-embedding-bge-large-en-v1.5} \\
SAG-Mem / Lite & Retrieval engine & SAG2 \texttt{0.3.0} \\
SAG-Mem & Evidence Evaluator & \texttt{typesafe/jev-1.14-20261014} \\
Ablation (\Cref{sec:ablation}) & Evidence Evaluator & Zev-1 (\texttt{knowline-35b-a3b-fp8}) \\
\bottomrule
\end{tabular}
\end{table}

\subsection{Overall Effectiveness}
\label{sec:effectiveness}

\mypara{Overall Comparison under Qwen}
\Cref{tab:main_results} reports the end-to-end answer accuracy of all evaluated systems under the Qwen configuration. Overall, both \systemname and \systemnamelite achieve strong performance without relying on LLM-based memory extraction or explicit graph construction, from which we draw three key conclusions:
\begin{enumerate}[label=(\arabic*), leftmargin=16pt, itemsep=2pt, topsep=2pt, parsep=0pt, partopsep=0pt]
    \item \systemname achieves the highest overall accuracy across both benchmarks, while \systemnamelite consistently ranks among the top-performing methods. Both variants substantially outperform all memory-based baselines, demonstrating the effectiveness of preserving complete source messages and discovering connections at retrieval time.
    \item Compared with standard RAG baselines operating over raw conversation history, both \systemname and \systemnamelite consistently outperform BM25-only and Dense RAG, confirming that evidence-guided lexical expansion and candidate scoring provide gains beyond single-pass retrieval.
    \item Relative to Full Context, \systemname maintains an advantage on LoCoMo and delivers a substantial improvement on LongMemEval, where lengthy multi-session histories degrade full-context reading. Although \systemnamelite trails Full Context on LoCoMo, it outperforms all other memory-based and RAG baselines there and surpasses Full Context on LongMemEval by a wide margin.
\end{enumerate}

\mypara{Category-Level Performance}
To examine how different retrieval and rendering components contribute across task types, we further analyze the category-level breakdown in \Cref{tab:main_results} under the Qwen configuration. Overall, the advantages of \systemname and \systemnamelite are most pronounced on categories requiring cross-session evidence connection and temporal grounding:
\begin{enumerate}[label=(\arabic*), leftmargin=16pt, itemsep=2pt, topsep=2pt, parsep=0pt, partopsep=0pt]
    \item On LoCoMo, \systemname establishes a clear lead on temporal reasoning over all baselines, reflecting the benefit of post-retrieval time clarification. On multi-hop questions, \systemname significantly outperforms all memory-based and standard RAG baselines and closely approaches Full Context; we diagnose this remaining multi-hop gap in \Cref{sec:retrieval}. On adversarial questions, retrieval-based systems, including both \systemname and \systemnamelite, consistently outperform Full Context in abstention.
    \item On LongMemEval, \systemname demonstrates superior performance on multi-session, temporal, and single-session user questions, while both variants achieve perfect accuracy on single-session assistant queries. Meanwhile, \systemnamelite attains the strongest performance on knowledge updates, surpassing both memory-based baselines and \systemname, indicating that fixed-query embedding scoring over lexically expanded candidate pools is particularly effective for locating updated factual statements.
\end{enumerate}

\mypara{Cross-Model Consistency under DeepSeek}
To examine whether selective retrieval remains beneficial under a stronger LLM configuration, we conduct a focused cross-model comparison of \systemname and \systemnamelite against the Full Context reference under DeepSeek, instead of re-evaluating the full baseline suite (whose cost is dominated by write-time memory extraction). Detailed and cross-model results are reported in \Cref{tab:main_results} and \Cref{tab:cross_model}, where both the Reader and Judge switch to the indicated LLM configuration:
\begin{enumerate}[label=(\arabic*), leftmargin=16pt, itemsep=2pt, topsep=2pt, parsep=0pt, partopsep=0pt]
    \item \systemname again attains the highest overall score on both benchmarks, edging out Full Context by a narrow margin on LoCoMo while maintaining a wide lead on LongMemEval. Under the DeepSeek configuration, \systemname slightly surpasses Full Context on LoCoMo multi-hop questions and retains clear advantages on temporal and adversarial categories. On LongMemEval, both \systemname and \systemnamelite continue to outperform Full Context by a wide margin across most categories.
    \item Moving from the Qwen configuration to DeepSeek yields higher overall scores for all three evaluated methods, reflecting the joint change in both the Reader and Judge models. Full Context records the largest gain on LongMemEval and also improves more than \systemname on LoCoMo, whereas \systemnamelite sees the largest increase on LoCoMo. Consequently, the relative lead of \systemname over Full Context narrows under DeepSeek, becoming a small edge on LoCoMo. Even so, both \systemname and \systemnamelite maintain a substantial advantage over Full Context on the longer multi-session histories of LongMemEval.
\end{enumerate}

\begin{table}[t]
\centering
\caption{End-to-end answer accuracy (\%) on LoCoMo and LongMemEval. \textbf{Bold} and \underline{underline} denote the highest and second-highest scores within each LLM configuration (when ties occur for first place, no second-highest score is underlined). Questions whose API requests failed are excluded from the denominator. Ove.: overall; M-H: multi-hop; O-D: open-domain; S-H: single-hop; Temp.: temporal; Adv.: adversarial; M-S: multi-session; K-U: knowledge update; S-P/S-A/S-U: single-session preference/assistant/user.}
\label{tab:main_results}
\setlength{\tabcolsep}{3.5pt}
\resizebox{\textwidth}{!}{%
\begin{tabular}{lcccccc|ccccccc}
\toprule
\multirow{2}{*}{\textbf{Method}} & \multicolumn{6}{c|}{\textbf{LoCoMo}} & \multicolumn{7}{c}{\textbf{LongMemEval}} \\
\cmidrule(lr){2-7} \cmidrule(lr){8-14}
& \textbf{Ove.} & \textbf{M-H} & \textbf{O-D} & \textbf{S-H} & \textbf{Temp.} & \textbf{Adv.} & \textbf{Ove.} & \textbf{M-S} & \textbf{K-U} & \textbf{Temp.} & \textbf{S-P} & \textbf{S-A} & \textbf{S-U} \\
\midrule
\multicolumn{14}{l}{\textit{LLM configuration: \texttt{Qwen3.6-35B-A3B-FP8}}} \\
Full Context   & \underline{80.46} & \textbf{78.37} & \textbf{42.71} & \textbf{90.13} & 64.85 & 82.96 & 68.88 & 54.13 & 69.23 & 65.15 & 37.93 & \textbf{100.00} & 91.43 \\
Zep            & 58.25 & 46.09 & 32.29 & 53.50 & 42.99 & 91.47 & 57.77 & 42.10 & 61.04 & 54.14 & 50.00 & 67.86 & 85.71 \\
TDAM           & 53.67 & 45.39 & 29.16 & 62.66 & 33.64 & 61.66 & 54.70 & 40.60 & 71.80 & 34.60 & 16.70 & 91.10 & 87.10 \\
Mem0           & 73.91 & 59.92 & 34.37 & 74.31 & \underline{71.65} & \underline{92.15} & 68.80 & 55.64 & 74.36 & 63.16 & \textbf{63.33} & 76.79 & 94.29 \\
MemOS          & 65.06 & 54.61 & 31.25 & 64.09 & 54.21 & 88.57 & 69.00 & 57.89 & \underline{79.49} & 70.68 & \underline{60.00} & 53.57 & 91.43 \\
BM25-only      & 72.60 & 48.20 & 27.10 & 82.50 & 54.80 & 91.90 & 67.60 & 46.60 & 74.40 & 64.70 & 40.00 & \textbf{100.00} & 91.43 \\
Dense RAG      & 76.20 & 67.00 & 35.40 & 82.00 & 57.90 & \textbf{93.00} & 73.00 & \underline{65.40} & 76.90 & 62.40 & 53.30 & \textbf{100.00} & 90.00 \\
\systemnamelite & 77.09 & 64.89 & 37.50 & 81.69 & 68.54 & 90.81 & \underline{77.20} & 61.65 & \textbf{85.90} & \underline{72.18} & \underline{60.00} & \textbf{100.00} & \underline{95.71} \\
\systemname    & \textbf{83.08} & \underline{75.89} & \underline{40.62} & \underline{88.35} & \textbf{79.44} & 89.46 & \textbf{80.20} & \textbf{72.93} & 78.21 & \textbf{76.69} & 53.33 & \textbf{100.00} & \textbf{98.57} \\
\midrule
\multicolumn{14}{l}{\textit{LLM configuration: \texttt{DeepSeek-V4.1-Flash-1017}}} \\
Full Context   & \underline{83.43} & \underline{79.08} & \textbf{51.04} & \textbf{92.39} & \underline{76.95} & 80.94 & 74.40 & 60.90 & \textbf{87.18} & 67.67 & 30.00 & \textbf{100.00} & 97.14 \\
\systemnamelite & 81.22 & 75.18 & \underline{48.96} & 87.40 & 76.32 & \textbf{83.86} & \underline{81.60} & \underline{69.17} & \underline{85.90} & \underline{78.95} & \textbf{63.33} & \textbf{100.00} & \underline{98.57} \\
\systemname    & \textbf{83.99} & \textbf{79.79} & 44.79 & \underline{90.61} & \textbf{82.24} & \textbf{83.86} & \textbf{82.60} & \textbf{72.93} & 83.33 & \textbf{79.70} & \textbf{63.33} & \textbf{100.00} & \textbf{100.00} \\
\bottomrule
\end{tabular}%
}
\end{table}

\begin{table}[!htbp]
\centering
\small
\caption{Cross-model comparison between \texttt{Qwen3.6-35B-A3B-FP8} and \texttt{DeepSeek-V4.1-Flash-1017}. Both the Reader and Judge use the indicated LLM configuration. Scores are overall accuracy (\%); $\Delta$ denotes DeepSeek minus Qwen in percentage points.}
\label{tab:cross_model}
\begin{tabular}{lccc|ccc}
\toprule
\multirow{2}{*}{\textbf{Method}} & \multicolumn{3}{c|}{\textbf{LoCoMo}} & \multicolumn{3}{c}{\textbf{LongMemEval}} \\
\cmidrule(lr){2-4} \cmidrule(lr){5-7}
& \textbf{Qwen} & \textbf{DeepSeek} & \textbf{$\Delta$} & \textbf{Qwen} & \textbf{DeepSeek} & \textbf{$\Delta$} \\
\midrule
Full Context    & 80.46 & 83.43 & +2.97 & 68.88 & 74.40 & +5.52 \\
\systemnamelite & 77.09 & 81.22 & +4.13 & 77.20 & 81.60 & +4.40 \\
\systemname     & 83.08 & 83.99 & +0.91 & 80.20 & 82.60 & +2.40 \\
\bottomrule
\end{tabular}
\end{table}

\subsection{System Efficiency: Cost and Latency}
\label{sec:efficiency}

\mypara{Monetary Cost and Token Breakdown}
\Cref{fig:cost_accuracy} summarizes the accuracy--cost trade-off, and \Cref{tab:cost} reports the total monetary cost and stage-level token breakdown of all evaluated systems under the Qwen configuration, separating write-time extraction and embedding from query-time retrieval and Reader generation. Overall, eliminating write-time LLM extraction yields decisive cost advantages as conversation histories scale, from which we draw three key observations:
\begin{enumerate}[label=(\arabic*), leftmargin=16pt, itemsep=2pt, topsep=2pt, parsep=0pt, partopsep=0pt]
    \item Memory-based baselines incur substantial write-time LLM extraction costs, which dominate their total expenditure on LongMemEval where hundreds of independent histories must be indexed. In contrast, both \systemname and \systemnamelite require zero extraction LLM tokens, keeping their total cost one to two orders of magnitude lower than memory-based systems on LongMemEval. On LoCoMo, where many queries share a small set of histories, write-time extraction amortizes more easily; \systemnamelite remains cheaper than all memory-based baselines, whereas \systemname allocates its budget to query-time JEV scoring and falls between the lighter memory baselines (Mem0 and MemOS) and the heavier ones (TDAM and Zep).
    \item \systemnamelite achieves the lowest total cost among Full Context and all memory-based baselines across both benchmarks, remaining comparable in cost to standard RAG baselines while delivering higher accuracy. Because it relies on lightweight lexical expansion and precomputed chunk embeddings, its primary overhead on LongMemEval lies in one-time embedding construction, which amortizes cleanly over subsequent queries. By comparison, \systemname eliminates embedding costs entirely and allocates most of its budget to query-time JEV evidence scoring, resulting in higher cost than \systemnamelite on LoCoMo yet still substantially undercutting Full Context and all memory-based baselines on LongMemEval.
    \item Selective retrieval reduces Reader token usage relative to Full Context on both benchmarks. On LongMemEval, this reduction yields substantial total cost savings for both variants. On LoCoMo, the estimated total cost of \systemname is 54\% lower than Full Context under the same uncached Reader rates, including query-time JEV scoring.
\end{enumerate}

\begin{table}[!htbp]
\centering
\small
\caption{Operational cost (USD) and token breakdown under the Qwen configuration. Extract.\ LLM and Search/Read LLM are reported in millions of tokens (M). Emb.\ denotes embedding cost (USD). \systemname uses 53.31M (LoCoMo) and 36.63M (LongMemEval) JEV input and output tokens. JEV costs are included in the total. Bold and underline indicate the lowest and second-lowest total cost on each benchmark.}
\label{tab:cost}
\resizebox{\textwidth}{!}{%
\begin{tabular}{lcccc|cccc}
\toprule
\multirow{2}{*}{\textbf{Method}} & \multicolumn{4}{c|}{\textbf{LoCoMo}} & \multicolumn{4}{c}{\textbf{LongMemEval}} \\
\cmidrule(lr){2-5} \cmidrule(lr){6-9}
 & \textbf{Total (\$)} & \textbf{Extract.\ (M)} & \textbf{Search/Read (M)} & \textbf{Emb.\ (\$)} & \textbf{Total (\$)} & \textbf{Extract.\ (M)} & \textbf{Search/Read (M)} & \textbf{Emb.\ (\$)} \\
\midrule
Full Context    & 6.20 & --  & 61.66 & --     & 9.80   & --    & 59.47 & --    \\
Zep             & 9.18 & 48.69 & 4.91  & 0.0024 & 687.49 & 4257.52 & 1.08  & 0.372 \\
TDAM            & 3.53 & 26.59 & 3.14  & 0.0030 & 253.41 & 1353.21 & 3.33  & 0.491 \\
Mem0            & 0.71 & 7.00  & 1.23  & 0.0041 & 189.86 & 3659.63 & 0.41  & 0.772 \\
MemOS           & 0.95 & 1.26  & 3.64  & 0.0028 & 40.07  & 233.43  & 1.08  & 0.481 \\
BM25-only       & \textbf{0.67} & --  & 6.38  & --     & \textbf{0.40}   & --
& 3.70  & --    \\
Dense RAG       & 0.71 &  -- & 6.62  & 0.0032 & \underline{0.79}   & --
& 2.18  & 0.539 \\
\systemnamelite & \underline{0.70} & --  & 6.45  & 0.0036 & 0.95 & --    & 3.33  & 0.583 \\
\systemname     & 2.86 & --  & 7.36  & --      & 1.91 & --    & 3.81  & --     \\
\bottomrule
\end{tabular}%
}
\end{table}

\mypara{End-to-End Query Latency}
\Cref{tab:latency} breaks down average query-time latency into retrieval time and Reader answer generation time over prebuilt indexes:
\begin{enumerate}[label=(\arabic*), leftmargin=16pt, itemsep=2pt, topsep=2pt, parsep=0pt, partopsep=0pt]
    \item \systemnamelite achieves low end-to-end latency, responding faster than Full Context, Zep, MemOS, and both standard RAG baselines, aided by faster Reader generation than Full Context.
    \item \systemname incurs higher end-to-end latency than \systemnamelite and the baselines, driven primarily by query-time JEV evidence scoring during retrieval alongside longer Reader generation time.
\end{enumerate}

\begin{table}[!htbp]
\centering
\small
\caption{End-to-end query latency (seconds) under the Qwen on LoCoMo, decomposed into query-time retrieval and Reader answer generation over prebuilt indexes. Bold and underline indicate the lowest and second-lowest values in each column.}
\label{tab:latency}
\begin{tabular}{lccc}
\toprule
\textbf{Method} & \textbf{Retrieval (s)} & \textbf{Generation (s)} & \textbf{Total (s)} \\
\midrule
Full Context    & \textbf{0.0012}    & 1.0429             & 1.0441             \\
Zep             & 0.1908             & 0.7748             & 0.9656             \\
TDAM            & 0.0616 & 0.7753             & \underline{0.8369} \\
Mem0            & 0.0890             & \textbf{0.5097}    & \textbf{0.5987}    \\
MemOS           & 0.2450             & 0.7947             & 1.0397             \\
BM25-only       & \underline{0.0085}             & 1.0238             & 1.0323             \\
Dense RAG             & 0.0534             & 1.2213             & 1.2747             \\
\systemnamelite & 0.2528             & \underline{0.6528} & 0.9056             \\
\systemname     & 2.0590             & 1.2230             & 3.2820             \\
\bottomrule
\end{tabular}
\end{table}

\subsection{Component Ablations}
\label{sec:ablation}

\mypara{Evidence-Guided Lexical Expansion}
\Cref{tab:ablation_expansion} compares the full \systemname retrieval pipeline ($H=1$) against a no-expansion control ($H=0$) under the Qwen configuration, which retains initial BM25 candidates, initial JEV scoring, and post-retrieval time clarification while removing the lexical expansion round and rescoring with evidence.
Removing the expansion round leads to a pronounced accuracy drop on both LoCoMo and LongMemEval, confirming that initial-query matching alone misses complementary facts needed for cross-session questions.
This consistent gain across both benchmarks validates that using selected source chunks to guide a lexical expansion round and rescoring with evidence is a primary contributor to the effectiveness of \systemname.

\begin{table}[!htbp]
\centering
\small
\caption{Evidence-guided lexical expansion ablation for \systemname under the Qwen configuration. Scores are overall accuracy (\%); $\Delta$ denotes No exp.\ ($H=0$) minus Main ($H=1$) in percentage points.}
\label{tab:ablation_expansion}
\begin{tabular}{lcccc}
\toprule
\textbf{Dataset} & \textbf{$N$} & \textbf{Main ($H=1$)} & \textbf{No exp.\ ($H=0$)} & \textbf{$\Delta$} \\
\midrule
LoCoMo      & 1,986 & \textbf{83.08} & 78.15 & $-4.93$ \\
LongMemEval & 500   & \textbf{80.20} & 75.20 & $-5.00$ \\
\bottomrule
\end{tabular}
\end{table}

\mypara{Evidence Evaluator Decoupling}
To test whether \systemname relies solely on JEV, we replace JEV with Zev-1 while keeping the rest of the pipeline unchanged (\Cref{tab:ablation_decision}).
Under the Qwen configuration, Zev-1 slightly outperforms JEV on LoCoMo and trails it on LongMemEval, yet still outperforms all memory-based and standard RAG baselines in \Cref{tab:main_results} by a wide margin.

\begin{table}[!htbp]
\centering
\small
\caption{Evidence Evaluator ablation for \systemname under the Qwen configuration, comparing JEV against Zev-1. Scores are overall accuracy (\%); bold indicates the higher score on each benchmark.}
\label{tab:ablation_decision}
\begin{tabular}{lcc}
\toprule
\textbf{Evidence Evaluator} & \textbf{LoCoMo} & \textbf{LongMemEval} \\
\midrule
JEV   & 83.08          & \textbf{80.20} \\
Zev-1 & \textbf{83.50} & 77.60          \\
\bottomrule
\end{tabular}
\end{table}

\mypara{Post-Retrieval Time Clarification}
\Cref{tab:ablation_temporal} evaluates post-retrieval time clarification on the temporal subsets of both benchmarks under the Qwen configuration, keeping the retrieved chunks fixed.
We compare three rendering settings: \textbf{Main} combines normalized date headers with inline date annotations for supported relative-time expressions, \textbf{Raw dates} keeps the original message headers without inline annotations, and \textbf{Date only} normalizes the headers while omitting inline annotations.
\begin{enumerate}[label=(\arabic*), leftmargin=16pt, itemsep=2pt, topsep=2pt, parsep=0pt, partopsep=0pt]
    \item On LoCoMo, \textbf{Main} delivers substantial accuracy gains over both \textbf{Raw dates} and \textbf{Date only} for \systemname and \systemnamelite. In contrast, \textbf{Date only} offers no advantage over \textbf{Raw dates}, indicating that explicit inline date annotations for relative-time expressions are essential for the Qwen Reader.
    \item On LongMemEval, the effect is smaller: \textbf{Main} improves over both controls for \systemnamelite, whereas for \systemname it matches \textbf{Raw dates} and edges out \textbf{Date only}. Across both benchmarks, deterministic post-retrieval time clarification provides consistent or positive gains without modifying stored indexes or incurring additional LLM calls.
\end{enumerate}

\begin{table}[!htbp]
\centering
\small
\caption{Post-retrieval temporal rendering ablation on temporal subsets under the Qwen configuration with retrieved chunks fixed. Scores are accuracy (\%); bold and underline indicate the highest and second-highest values in each row (ties for first are bolded without underlining).}
\label{tab:ablation_temporal}
\begin{tabular}{lccc}
\toprule
\textbf{Dataset \& Method} & \textbf{Main} & \textbf{Raw dates} & \textbf{Date only} \\
\midrule
LoCoMo ($N=321$) --- \systemname         & \textbf{79.44} & \underline{63.55} & 61.37 \\
LoCoMo ($N=321$) --- \systemnamelite     & \textbf{68.54} & \underline{56.39} & 56.07 \\
LongMemEval ($N=133$) --- \systemname     & \textbf{76.69} & \textbf{76.69}    & 75.19 \\
LongMemEval ($N=133$) --- \systemnamelite & \textbf{72.18} & 69.17             & \underline{70.68} \\
\bottomrule
\end{tabular}
\end{table}

\subsection{Retrieval Diagnostics and Multi-Hop Bottleneck Analysis}
\label{sec:retrieval}

\mypara{Retrieval-Layer Recall}
\Cref{tab:retrieval_recall} reports session-level and turn-level Recall@5/10/20 for \systemname and \systemnamelite across both benchmarks, decoupling retrieval coverage from downstream Reader generation.
\systemname consistently achieves high session-level and turn-level recall on both LoCoMo and LongMemEval, capturing the vast majority of supporting evidence within the top-5 chunks and approaching saturation by $K=20$.
By comparison, \systemnamelite trails \systemname at tighter cutoffs, especially on turn-level recall, yet substantially narrows the gap at $K=20$.
This divergence reflects the benefit of JEV evidence scoring in concentrating gold evidence into top ranks, while explaining why \systemnamelite remains competitive in end-to-end accuracy once twenty chunks are supplied to the Reader.

\begin{table}[!htbp]
\centering
\small
\caption{Retrieval-layer session-level and turn-level Recall@5/10/20 (\%) for \systemname and \systemnamelite on LoCoMo ($N=1{,}973$ annotated questions) and LongMemEval ($N=500$).}
\label{tab:retrieval_recall}
\begin{tabular}{llccc}
\toprule
\textbf{Method} & \textbf{Dataset} & \textbf{$N$} & \textbf{Session R@5 / R@10 / R@20} & \textbf{Turn R@5 / R@10 / R@20} \\
\midrule
\systemname     & LoCoMo      & 1,973 & 94.44 / 96.56 / 97.85 & 90.60 / 93.09 / 94.15 \\
\systemnamelite & LoCoMo      & 1,973 & 79.45 / 87.91 / 94.14 & 66.33 / 75.99 / 84.62 \\
\systemname     & LongMemEval & 500   & 96.19 / 97.66 / 98.40 & 91.11 / 94.88 / 95.72 \\
\systemnamelite & LongMemEval & 500   & 89.98 / 96.22 / 98.22 & 77.56 / 90.25 / 94.02 \\
\bottomrule
\end{tabular}
\end{table}

\mypara{Multi-Hop Bottleneck Analysis}
To diagnose the remaining multi-hop gap relative to Full Context under Qwen in \Cref{sec:effectiveness}, we perform a targeted retrieval-budget sweep and error attribution on the LoCoMo multi-hop subset using \systemname (\Cref{tab:multihop_sweep} and \Cref{tab:multihop_attribution}).
The diagnostic results show that this gap does not stem from a failure to locate gold evidence, but instead reflects an interaction between the returned chunk budget $K$ and Reader-side multi-hop reasoning.
At the default setting ($K=20, H=1$), message and session coverage are already high, and expanding $K$ or adding a second lexical expansion round ($H=2$) produces negligible changes in aggregate coverage.
Nevertheless, enlarging $K$ from 20 to 40 yields a clear accuracy gain that closes and reverses the gap to Full Context, after which accuracy saturates at $K=60$.
This pattern aligns with the sensitivity to $K$ observed on LongMemEval in \Cref{sec:sensitivity}: even when most annotated gold turns are already retrieved, additional chunks supply surrounding conversational context and remaining partial hops that help the Qwen Reader link scattered facts, at the cost of a larger Reader context.
Adding a second expansion round ($H=2$) also improves accuracy modestly while increasing query-time JEV scoring overhead.
As detailed in \Cref{tab:multihop_attribution}, complete retrieval misses account for only a small fraction of failures under both $H=1$ and $H=2$, while the vast majority of errors still occur on questions with gold evidence already present in the Reader context.
Combined with the DeepSeek results in \Cref{sec:effectiveness}, where \systemname surpasses Full Context even at $K=20$, these diagnostics indicate that default multi-hop accuracy is jointly bounded by the $K=20$ context trade-off and Reader reasoning capacity.

\begin{table}[!htbp]
\centering
\small
\caption{Multi-hop retrieval-budget sweep for \systemname on the LoCoMo multi-hop subset ($N=282$) under the Qwen configuration. Bold indicates the highest value in each metric column. Each configuration is evaluated in an independent end-to-end run; minor non-monotonic fluctuations in coverage reflect small score variations across separate JEV API calls.}
\label{tab:multihop_sweep}
\begin{tabular}{lccccccc}
\toprule
\textbf{Config} & \textbf{$B$} & \textbf{$H$} & \textbf{$K$} & \textbf{Correct / Total} & \textbf{Acc.\ (\%)} & \textbf{Msg.\ Cov.\ (\%)} & \textbf{Sess.\ Cov.\ (\%)} \\
\midrule
\texttt{k20\_h1} & 50 & 1 & 20 & 214 / 282 & 75.9 & 94.6 & 96.8 \\
\texttt{k40\_h1} & 50 & 1 & 40 & 226 / 282 & 80.1 & 94.3 & 97.0 \\
\texttt{k60\_h1} & 60 & 1 & 60 & 227 / 282 & \textbf{80.5} & \textbf{95.8} & \textbf{97.2} \\
\texttt{k20\_h2} & 50 & 2 & 20 & 220 / 282 & 78.0 & 95.1 & 96.9 \\
\bottomrule
\end{tabular}
\end{table}

\begin{table}[!htbp]
\centering
\small
\caption{Error attribution breakdown on the LoCoMo multi-hop subset ($N=282$) for \systemname under the Qwen configuration. \textbf{Complete Miss} counts incorrect answers for which none of the annotated gold evidence turns appear in the returned $K=20$ chunks; \textbf{Evidence Present, Wrong} counts incorrect answers for which at least one gold evidence turn is retrieved (covering both full and partial gold-turn coverage).}
\label{tab:multihop_attribution}
\begin{tabular}{lccc}
\toprule
\textbf{Configuration} & \textbf{Correct} & \textbf{Complete Miss} & \textbf{Evidence Present, Wrong} \\
\midrule
$K=20, H=1$ & 214 & 4 & 64 \\
$K=20, H=2$ & 220 & 1 & 61 \\
\bottomrule
\end{tabular}
\end{table}

\subsection{Hyperparameter Sensitivity Analysis}
\label{sec:sensitivity}

To examine how retrieval budgets affect performance without incurring full-benchmark JEV evaluation costs across all parameter combinations, we conduct a sensitivity sweep for \systemname under the Qwen configuration on a category-stratified 125-question subset of LongMemEval, complementing the LoCoMo multi-hop budget sweep in \Cref{sec:retrieval}. We vary three core retrieval hyperparameters around the fixed default configuration (\texttt{default}: $B=50, H=1, K=20$): the per-round candidate pool size $B$, the maximum number of lexical expansion rounds $H$, and the number of chunks passed to the Reader $K$ (\Cref{tab:hyperparam_sweep}). Overall, subset accuracy responds noticeably to retrieval budget adjustments across all three dimensions, while the fixed default setting provides a balanced trade-off between accuracy and query-time overhead:
\begin{enumerate}[label=(\arabic*), leftmargin=16pt, itemsep=2pt, topsep=2pt, parsep=0pt, partopsep=0pt]
    \item Among the three parameters, accuracy is most sensitive to the returned chunk budget $K$. Reducing $K$ below the default lowers accuracy, whereas doubling $K$ yields the largest single-parameter gain on the subset at the cost of doubling the Reader input context.
    \item Adjusting the candidate pool size $B$ and expansion depth $H$ produces asymmetric effects. Shrinking $B$ leaves subset accuracy unchanged, whereas disabling lexical expansion ($H=0$) degrades accuracy, consistent with the full-benchmark ablation in \Cref{sec:ablation}. Doubling either $B$ or $H$ yields identical moderate gains on the subset, though both increase query-time JEV scoring overhead.
\end{enumerate}

\begin{table}[!htbp]
\centering
\small
\caption{Hyperparameter sensitivity sweep for \systemname on a category-stratified LongMemEval subset ($N=125$) under the Qwen configuration. \texttt{default} denotes the fixed default configuration ($B=50, H=1, K=20$). Bold indicates the highest accuracy.}
\label{tab:hyperparam_sweep}
\begin{tabular}{lccccc}
\toprule
\textbf{Config} & \textbf{$B$} & \textbf{$H$} & \textbf{$K$} & \textbf{Correct / Total} & \textbf{Acc.\ (\%)} \\
\midrule
\texttt{default} & 50  & 1 & 20 & 94 / 125  & 75.2 \\
\texttt{k10}     & 50  & 1 & 10 & 90 / 125  & 72.0 \\
\texttt{k40}     & 50  & 1 & 40 & 100 / 125 & \textbf{80.0} \\
\texttt{b20}     & 20  & 1 & 20 & 94 / 125  & 75.2 \\
\texttt{b100}    & 100 & 1 & 20 & 98 / 125  & 78.4 \\
\texttt{h0}      & 50  & 0 & 20 & 91 / 125  & 72.8 \\
\texttt{h2}      & 50  & 2 & 20 & 98 / 125  & 78.4 \\
\bottomrule
\end{tabular}
\end{table}

\section{Related Work}
\mypara{Memory Representation and Organization}
External conversational memories differ in the granularity of stored content and the structures used to organize it~\citep{hu2025memory}.
Summarization and reflection produce compact descriptions of past interactions or experiences, including event summaries, user profiles, and higher-level insights~\citep{zhong2024memorybank,park2023generative,tan2025prospect}.
Other approaches represent salient information as persistent facts or self-contained discourse units, preserving details for fine-grained retrieval~\citep{chhikara2025mem0,zhou2025simple}.
Hierarchical designs organize memory into tiers and manage information transfer or consolidation across them~\citep{packer2023memgpt,fang2026lightmem}.
Temporal knowledge graphs and linked-note structures maintain associations among memories and their context~\citep{rasmussen2025zep,xu2026mem}.
Our framework preserves complete dialogue messages and their metadata in chunks, avoiding LLM-based memory rewriting and persistent semantic relation construction.

\mypara{Memory Retrieval Strategy}
Retrieval strategies seek supporting evidence beyond the matches returned for the original question.
Relevance feedback and adaptive re-ranking use initial results or highly scored documents to guide subsequent search~\citep{lavrenko2017relevance,macavaney2022adaptive}.
Generated reasoning steps, draft answers, and predicted continuations provide additional retrieval cues~\citep{trivedi2023interleaving,qian2025memorag,jiang2023active}.
Graph-based retrieval exploits entity and passage connectivity to recover evidence distributed across documents~\citep{gutierrez2024hipporag,gutierrez2025rag}.
SAG~\citep{wu2026sag} connects source passages at query time through shared entities in an event--entity index.
SAG2~\citep{wu2026sag2} combines BM25 retrieval with decision-model evidence scoring and lexical expansion from selected passages, without offline embedding computation or LLM-based relation extraction.
Our work adapts SAG2's evidence-guided lexical expansion to conversational memory and evaluates JEV-based and embedding-based scoring configurations.

\section{Conclusion}
\label{sec：conclusion}
We presented \systemname and \systemnamelite, two configurations of a conversational memory framework that preserves complete dialogue messages and discovers complementary evidence through seed-based lexical expansion.
\systemname combines model-free memory construction with JEV-based evidence evaluation, while \systemnamelite uses precomputed chunk embeddings for candidate scoring.
Both configurations append date-based interpretations of supported relative-time expressions while preserving the original text and temporal granularity.

On LoCoMo and LongMemEval, \systemname achieves overall scores of 83.08 and 80.20 with Qwen3.6-35B, and 83.99 and 82.60 with DeepSeek-V4.1-Flash, respectively, outperforming the Full Context baseline in all four settings.
\systemnamelite achieves corresponding scores of 77.09 and 77.20 with Qwen, and 81.22 and 81.60 with DeepSeek, while maintaining low query latency and lower operational cost than Full Context and all memory-based baselines.
These results support dialogue preservation and evidence-guided retrieval expansion as an effective design for long-term conversational memory across both evaluated LLM configurations.

\bibliographystyle{unsrtnat}
\bibliography{references} 

\appendix

\section{JEV Request Packing}
\label{app:packing}

A pool is scored through one or more JEV requests, and the set $V_t(i)$ in Eqs.~\eqref{eq:jev0} and~\eqref{eq:jev1} is determined by how the pool is packed into them.

\mypara{Pool Order}
The client walks the pool in its existing order.
For the initial pool $C_0$ this is BM25 order.
For the rescoring pool $U_1$ it is the five chunks of $P$, then the seeds, then the new candidates in retrieval order.

\mypara{Token Estimate}
The client accumulates a token estimate with its own \texttt{o200k\_base} counter, which is distinct from the BGE counts used for chunking in Section~\ref{sec:construction}.
It opens a new request once the estimate exceeds 64{,}000.

\mypara{Shared State}
The candidates of a request share one state, which holds the question and the explicit evidence list and, in stateful mode, also the passages of all candidates in the request and the common scoring instruction.
This state is never shared across requests.
In stateful mode, if it exceeds 32{,}000 tokens or the whole request exceeds 64{,}000, the current candidate list is halved and packed again.
Both thresholds are settings of the local client rather than limits of the model.

\mypara{Packing Determinism.}
Requests are sent concurrently and scores are written back by candidate identifier.
The client-side packing, and with it $V_t(i)$, is therefore fixed by the pool and its order and does not depend on the returned scores.
When a candidate or the evidence itself has to be divided, the candidate keeps the maximum of its partial scores, while the stored chunk stays whole.
This guarantees deterministic request construction on the client side; separate end-to-end runs calling the remote JEV API can still exhibit minor floating-point score fluctuations near seed-selection boundaries, accounting for small coverage variations across independent runs (\Cref{tab:multihop_sweep}).

\end{document}